\documentclass[12pt,a4paper]{article}
\usepackage[utf8]{inputenc}
\usepackage[vietnamese,english]{babel}
\usepackage[T5,T1]{fontenc}

\newcommand*\vn{\fontencoding{T5}\selectfont\selectlanguage{vietnamese}}
\usepackage{amsmath, amssymb, fullpage, graphics, amsthm}
\usepackage{caption, graphicx, fancyhdr, fancybox}
\usepackage{subcaption}
\newtheorem{theorem}{Theorem}

\newtheorem{statement}[theorem]{Statement}
\newtheorem{remark}{Remark}
\usepackage{color}
\usepackage{tabularx}
\usepackage[linesnumbered,lined,boxed,commentsnumbered]{algorithm2e}
\usepackage{multicol}
\usepackage{multirow}

\SetAlFnt{\small}
\SetAlCapFnt{\small}
\SetAlCapNameFnt{\small}
\SetKwComment{Comment}{/* }{ */}
\usepackage{hyperref}
\usepackage{cite}

\usepackage{pgf}
\usepackage[numbers,sort]{natbib}

\makeatletter
\fancypagestyle{footmark}{
  \ps@@fancy 
  \fancyfoot[C]{\footmark}
}

\makeatother
\begin{document}

\begin{center}
{\large\sf \textbf{CLUSTERING VIETNAMESE CONVERSATIONS FROM FACEBOOK PAGE TO BUILD TRAINING DATASET FOR CHATBOT}}
\vspace{2mm}

Trieu Hai Nguyen$^{1}$, Thi-Kim-Ngoan Pham$^{1}$, Thi-Hong-Minh Bui$^{1}$, \\ Thanh-Quynh-Chau Nguyen$^{1}$  \\[2mm]

${^1}$ Faculty of Information Technology, Nha Trang University, \\
02 Nguyen Dinh Chieu Street, Nha Trang City, Vietnam \\[2mm] 
e-mails: \{trieunh,ngoanptk,minhbth,chauntq\}@ntu.edu.vn

\end{center}
\begin{abstract}
The biggest challenge of building chatbots is training data. The required data must be realistic and large enough to train chatbots. We create a tool to get actual training data from Facebook messenger of a Facebook page. After text preprocessing steps, the newly obtained dataset generates FVnC and Sample dataset. We use the Retraining of BERT for Vietnamese (PhoBERT) to extract features of our text data. K-Means and DBSCAN clustering algorithms are used for clustering tasks based on output embeddings from PhoBERT$_{base}$. We apply V-measure score and Silhouette score to evaluate the performance of clustering algorithms. We also demonstrate the efficiency of PhoBERT compared to other models in feature extraction on the Sample dataset and wiki dataset. A GridSearch algorithm that combines both clustering evaluations is also proposed to find optimal parameters. Thanks to clustering such a number of conversations, we save a lot of time and effort to build data and storylines for training chatbot.
\let\thefootnote\relax\footnotetext{\small Corresponding author: Trieu Hai Nguyen \small (trieunh@ntu.edu.vn)
\\
\texttt{\small Preprint submitted to JJCIT} 
\hfill \textit{December 10, 2021}
}
\end{abstract}
Keywords: BERT, Clustering, Language models, Feature Extraction, Word Embeddings
\section{Introduction} \label{sec:1}
Chatbot is certainly not an unfamiliar name in the field of Natural Language Processing (NLP). Previously, traditional chatbot simply interacts with the user via predefined rules, which means the user is only allowed to enter  by these rules to get answers. However, NLP chatbot is not only a word recognition algorithm but it can also understand what the user is saying. It is one of the pioneering applications using Artificial Intelligence (AI) namely, NLP to help humans interact with machine like humans with humans via Virtual Assistant autoresponder. Currently, there are many parties developing NLP chatbot, which can be mentioned as Google's DialogFlow, Watson of IBM and Rasa. 

In the process of building NLP chatbots, all chatbots require real datasets for training bot. The training datasets can be large or small depending on the size and intelligence level of the chatbots. Raw training data can be collected from past conversations through social media, archived user chats, previous questions, email chains, or live telephone transcripts. But these data are messy, not in any structure or order, and come from various sources collected with huge amounts of raw data. Thus the first priority when constructing a chatbot is to transform those raw data into useful data for the purpose of training bot. 

In order to fit chatbot building orientation, that raw dataset needs to be divided into specific intents, which serves to build conversations to train a chatbot. There are many ways to process that raw dataset into specific intents (topics, conversations). The first method to be mentioned is using Supervised Learning task \citep{DBLP:journals/corr/abs-1805-05052} to classify intents. In particular, this method requires labeling for the input examples and then it predicts labels for remaining data in the raw dataset. In this case, label prediction corresponds to classified raw data into the intents that have been labeled previously. However, building and labeling manual intents on large datasets lead to big challenges for chatbot developers. With a simple raw dataset of about 8000 conversations, analyzing how many intents are created is a conundrum. 

Instead, we can approach the above problem by using the second method, which is clustering similar raw data together into corresponding intents. The advantage of this approach is that it uses Unsupervised Learning technical \citep{DBLP:journals/corr/abs-1805-05052}, which is only based on the features of the data to perform specific tasks such as Clustering. As expected, this method has a significant effect on the analysis of raw data. It saves us a lot of time and effort to make a training dataset for chatbot. There are many clustering algorithms such as K-Means, DBSCAN, BIRCH, Spectral clustering \citep{1017616,scikit-learn}. In this article, we use K-Means \citep{MacQueen1967} and DBSCAN \citep{ester1996densitybased} techniques to cluster our dataset and consider that clustering is a downstream NLP task. Each technique has its own advantages and disadvantages. K-Means algorithm is a  simple and fast implementation but it requires knowing the number of clusters to perform clustering whereas DBSCAN does not. Nevertheless, DBSCAN technique is more difficult to implement and requires finding the optimal parameters \citep{ester1996densitybased,Sander2004DensityBasedCI, Gaonkar2013AutoEpsDBSCAND}, which leads to drastically increased costs, especially for large datasets. Thus, we can combine the advantages of both techniques to serve the purpose of efficient clustering.

The input of clustering algorithms in particular and downstream NLP tasks in general is document embeddings extracted from the dataset. There are many ways to extract information from text datasets, for example, we can use traditional machine learning algorithm like TF-IDF \citep{salton1986introduction}, proposed word embedding models in recent years such as Word2Vec, GloVe \citep{mikolov, pennington}, FastText\citep{bojanowski2016enriching} or popular language models like GPT-2 \citep{radford2019language}, BERT model and its variations \citep{DBLP:journals/corr/abs-1810-04805, Sanh2019DistilBERTAD, DBLP:journals/corr/abs-2003-00744}. In this work, we use \textbf{BERT} (\textbf{B}idirectional \textbf{E}ncoder \textbf{R}epresentations from \textbf{T}ransformers), which is state-of-the-art embeddings \citep{DBLP:journals/corr/abs-1810-04805} to extract features of documents. Recently a clustering approach with the BERT model has been proposed by O. Gencoglu \citep{DBLP:journals/corr/abs-1901-00439}. As suggested in \citep{pugachev2021short} clustering techniques using pre-trained transformer language models are applied to short text clustering. The combination of word embeddings using BERT models and clustering algorithms to obtain topics was presented in \citep{sia2020tired}. In distinctive, PhoBERT is Pre-trained language models for Vietnamese, which is used to embed our Vietnamese dataset \citep{DBLP:journals/corr/abs-2003-00744}. V-measure score \citep{rosenberg-hirschberg-2007-v} and Silhouette score \citep{ROUSSEEUW198753} are used to evaluate the performance of a clustering algorithms as well as the feature extraction efficiency of the language models.

The aim of the present paper is to study and apply PhoBERT model to our Facebook Vietnamese conversations dataset, thereby deriving document embeddings in order to serve the clustering task. The combination of both  K-Means and DBSCAN clustering algorithms is proposed by us to achieve the best clustering results on the actual dataset. The finding of these data clusters allows us to simplify and accelerate the building of a training dataset for chatbot. In section \ref{sec:2}, we recall some theories of Transformer and BERT architecture proposed by Vaswani et al. in \citep{DBLP:journals/corr/VaswaniSPUJGKP17} and Devlin et al. in \citep{DBLP:journals/corr/abs-1810-04805} respectively. In section \ref{sec:3}, we offer an approach to apply PhoBERT to the clustering task from the idea of classification task \citep{DBLP:journals/corr/abs-2003-00744, sun2019finetune}. Next, we also recall clustering algorithms in machine learning and evaluation metrics for unsupervised learning algorithms in Section \ref{sec:4}. Some experiments on our Facebook Vietnamese conversations dataset (includes FVnC and Sample dataset) and wiki dataset such as searching optimal parameters, clustering performance evaluations as well as clustering results are considered in Section \ref{sec:exp}. In particular, we show that among the models that support Vietnamese,  PhoBERT's feature extraction efficiency is the best based on V-measure score. Ultimately, we give some conclusions in Section \ref{sec:6}. The code, datasets and pre-trained models are available at \url{https://github.com/trieuntu/conversation_clustering}.

\section{Related Work}\label{sec:2}
We provide some background knowledge about Transformer architecture, Pre-Trained Language Models especially BERT. From these theoretical constructs, we apply them to solve our NLP tasks.
\subsection{Transformer}
Transformer architecture was first introduced in paper ``Attention Is All You Need'' by \citep{DBLP:journals/corr/VaswaniSPUJGKP17}. At the time of launch, this architecture is a new breakthrough in the field of natural language processing and related tasks. Currently, when dealing with sequence-to-sequence models in NLP, the transformer is still one of the state-of-the-art (SOTA) types of model and completely replaces RNN/LSTM \citep{RNN}. Transformer architecture overcomes the disadvantages of RNN and its variations. For instance, it doesn't take advantage of GPU parallelism because it has to process input word by word sequentially into encoder/decoder, and the information is easily lost during propagation through hidden layers for long input sentences.  

Transformer architecture contains two parts, Encoder attention and Decoder attention. According to the original article of \citep{DBLP:journals/corr/VaswaniSPUJGKP17}, the encoder part has 6 layers, each of which has two sublayers, which are multi-head self-attention and fully connected feed-forward. Decoder part is similar to the encoder part but it adds a masked multi-head attention sublayer and the last layer of the encoder part will be passed to the multi-head attention sublayer in the decoder part. Note that, the input of both parts is the sum of positional encoding vector and word vector embedding.

The attention mechanism is the most important component of transformer architecture. Self-attention sublayer is an attention mechanism, which contains the weight sets of the model $W_q, W_k, W_v$ to be trained. The attention mechanism presents the relation of a word to all its related words in the sequence based on the adjustment of the above sets of weights. The product of the input embedding layer and $W_q, W_k, W_v$ is matrixes Query $Q$, Value $V$ and Key $K$. In order to calculate Attention vector of word $i$ to the rest of the words, Vaswani et al. \citep{DBLP:journals/corr/VaswaniSPUJGKP17} have given the formula
\[ Attention_i(Q_i, K_i, V_i ) = softmax \left(\frac{Q_i K_i^T}{\sqrt{d_{K_i}}}\right).V_i \]
where $d_K$ is the dimension of $K$. Each computed $Attention$ obtains a head-attention. We can compute the $Attention$ in parallel, which leads to the multi-head attention mechanism by concatenating head-attentions 
\[MultiHead(Q, K, V ) = Concatenate(head_1 , head_2, \ldots, head_n ).W_O\]
where $head_1$ corresponds to $Attention_1$. Matrix $W_O$ has the same number of columns as the input matrix. 

\subsection{Pre-Trained Language Models: BERT}
Training models from scratch on large datasets is impossible for most people. Thus using pre-trained models is an inevitable trend in the development of Artificial Intelligence. Taking into account the advantage of the weights that can be learned from trained models, we just need to fine-tune them to suit specific purposes. Formerly, pre-trained models in NLP have been mentioned in many studies \citep{mikolov,pennington,pmlr-v32-le14,fasttext2016,peters2018deep}. One of the great advantages of the transformer's architecture is that it allows the creation of NLP models trained, which can be reused in downstream NLP tasks. Some of the pre-trained language models based on transformer's architecture have achieved state-of-the-art results like BERT of \citep{DBLP:journals/corr/abs-1810-04805} from Google, GPT of Radford and Narasimhan \citep{Radford2018ImprovingLU} from Open AI and their variations. These new models can do things that the old models can't such as that it allows transfer learning in NLP with both low and high-level features. Transfer learning is a combination of reusing the architecture of pre-trained model and fine-tune parameters of the original layers to accommodate downstream tasks.

Specifically, BERT is an easily fine-tuned pre-train word embedding on a large unlabelled text corpus (unsupervised) which is trained based on Masked Language Model Task and Next Sentence Prediction Task. BERT's architecture is built only on the Encoder part of the Transformer. The input text before applying fine-tuning for Vietnamese in particular and other languages in general is a combination of Token Embeddings, Segment Embedding and Position Embeddings. If the input text consists of two or more sentences (pair-sequence), we must add token \texttt{[CLS]} at the beginning of the sentence and token \texttt{[SEP]} to separate the sentences. 

Masked Language Model task allows us to fine-tune word representations on any unsupervised text corpus. This task creates embeddings for the above Vietnamese dataset. The principle of operation of model training can be understood by predicting a missing word in the sequence instead of trying to predict the next word in the sequence itself. A missing word  is equivalent to \texttt{[MASK]} token.  We randomly mask 15\% of the total tokens in the sequence and predict these \texttt{[MASK]} tokens. Note that a missing word can be replaced by \texttt{[MASK]} token 80\% of the time, 10\% of the time for a random token and 10\% of the time for the unchanged token.

Next Sentence Prediction (NSP) is a binary classification task applied practically to the Question Answering (QA) task. NSP helps us to understand the relationship between sentences. The input of the model is a pair-sequence, which has been added tokens \texttt{[CLS]}, \texttt{[SEP]}. During model training, we select 50\% of the time of the second sentence, which is the next sentence of the first one and labeled as \texttt{IsNext}, while the remaining 50\% of the second sentence is randomly chosen from unrelated sentences in the dataset and labeled as \texttt{NotNext}.

There are many versions of BERT with different parameters on transformer architectures. The two most basic models are BERT$_{BASE}$ and BERT$_{LARGE}$. In essence, both models are the same but they are different in size. Specifically, according to Devlin et al. \citep{DBLP:journals/corr/abs-1810-04805} these models have the following sizes
\[
\text{BERT}_{BASE} (L=12, H=768, A=12, Total \text{ } Parameters=110M),
\]
\[
\text{BERT}_{LARGE} (L=24, H=1024,A=16, Total \text{ } Parameters=340M)
\]
where $L$ is the number of layers in the Encoder part of transformer architecture, $H$ is the hidden size and $A$ is the number of heads in multi-head self-attention.

\section{PhoBERT for Text Clustering}\label{sec:3}
PhoBERT is Pre-trained language models for Vietnamese proposed by Nguyen and Nguyen \citep{DBLP:journals/corr/abs-2003-00744}. At the time of launch, pre-trained PhoBERT models established state-of-the-art results in most tasks related to Vietnamese NLP. Although BERT can be applied to many tasks like Classification, Clustering, Dependency parsing, Sentiment analysis, Summarization text, Part-of-speech tagging, Question Answering, Named-entity recognition and Machine translation, in this work we only focus on clustering task to analyze our Vietnamese conversations dataset.

PhoBERT$_{base}$ and PhoBERT$_{large}$ are two versions of PhoBERT, whose architectures are similar to the BERT$_{BASE}$ and BERT$_{LARGE}$ above. PhoBERT uses RoBERTa , which is based on pytorch framework \citep{liu2019roberta} to retrain the BERT models on new 20GB pre-training Vietnamese dataset. Since PhoBERT architecture is based on RoBERTa, it only trains BERT model with Masked Language Model task. Another difference between PhoBERT and RoBERTa is fastBPE used to tokenize input sentences. Currently there are many methods to tokenize such as Word Level Tokenizer,  Multi-Word-Level Tokenizer, Character Level Tokenizer, Subword Units Level (BPE algorithm) Tokenizer but only BPE (\textbf{B}yte-\textbf{P}air \textbf{E}ncoding proposed by Sennrich et al. \citep{sennrich-etal-2016-neural}) achieves SOTA and is applied to most modern NLP models. 

BPE is a compression technique and is adapted for word segmentation tasks. Most words can be represented by subwords using the BPE method. It overcomes the disadvantages of Word and Character Tokenizers, for instance, words that do not appear in the dictionary can be represented in these subwords and the index length of sequence output is significantly shorter than Character Tokenizers. Code\footnote{Scripts are available at \url{https://github.com/rsennrich/subword-nmt}} of BPE algorithm to segment word into subword units was published by \citep{sennrich-etal-2016-neural}. For example, assume that the given Vietnamese vocabulary is
{\vn \footnotesize \begin{center}
vocab = \{'x i n h </w>': 10, 'đ ẹ p </w>': 20, 'x i n h \_ đ ẹ p </w>': 10, \\'x i n h \_ x ắ n </w>': 15, 'x ắ n </w>': 8\}
\end{center}}
Notice that unlike English, the Vietnamese language does not use white space to separate words because Vietnamese words can have more than one syllable. For illustration take a simple Vietnamese sentence {\vn ``\textit{Cô ấy rất xinh đẹp}''} (English version is ``\textit{She is very beautiful}''), which can be rewritten in the monosyllable form {\vn ``\textit{Cô\_ấy$_{\texttt{ She }}$ rất$_{\texttt{ very }}$ xinh\_đẹp$_{\texttt{ beautiful }}$}''}. Therefore, we can apply a Multi-Word-Level Tokenizer on the pre-training Vietnamese dataset before going into BPE. There are many toolkits to support word segmentation based on Multi-Word-Level Tokenization like RDRSegmenter from VnCoreNLP \citep{vu-etal-2018-vncorenlp}, pyvi \citep{pyvi}, underthesea \citep{undertheseanlp}. In the example above, tokens \texttt{</w>} are appended to the end of the words to mark the end of a word in Vietnamese vocabulary. After merging the most frequent pair at the \textit{9th} iteration, we obtain a new vocabulary as follows
{\vn \footnotesize \begin{center}
vocab$_{new}$=\{'xinh': 10, '</w>': 10, 'đẹp</w>': 30, 'xinh\_': 25, 'xắn</w>': 23\}
\end{center}}
\noindent It is clear that the word {\vn '\textit{xinh\_đẹp</w>}'} can be represented by subwords {\vn '\textit{xinh\_}'} and {\vn '\textit{đẹp</w>}'} from the above vocab$_{new}$. Especially, word {\vn '\textit{xinh\_xinh}'} (English meaning is \textit{pretty}) is out of vocabulary words, which can also be represented by the word pair {\vn '\textit{xinh\_}'} and {\vn '\textit{xinh}'}. 

\begin{figure}[!htb]
\centering \resizebox{0.5\textwidth}{!}{
\includegraphics[]{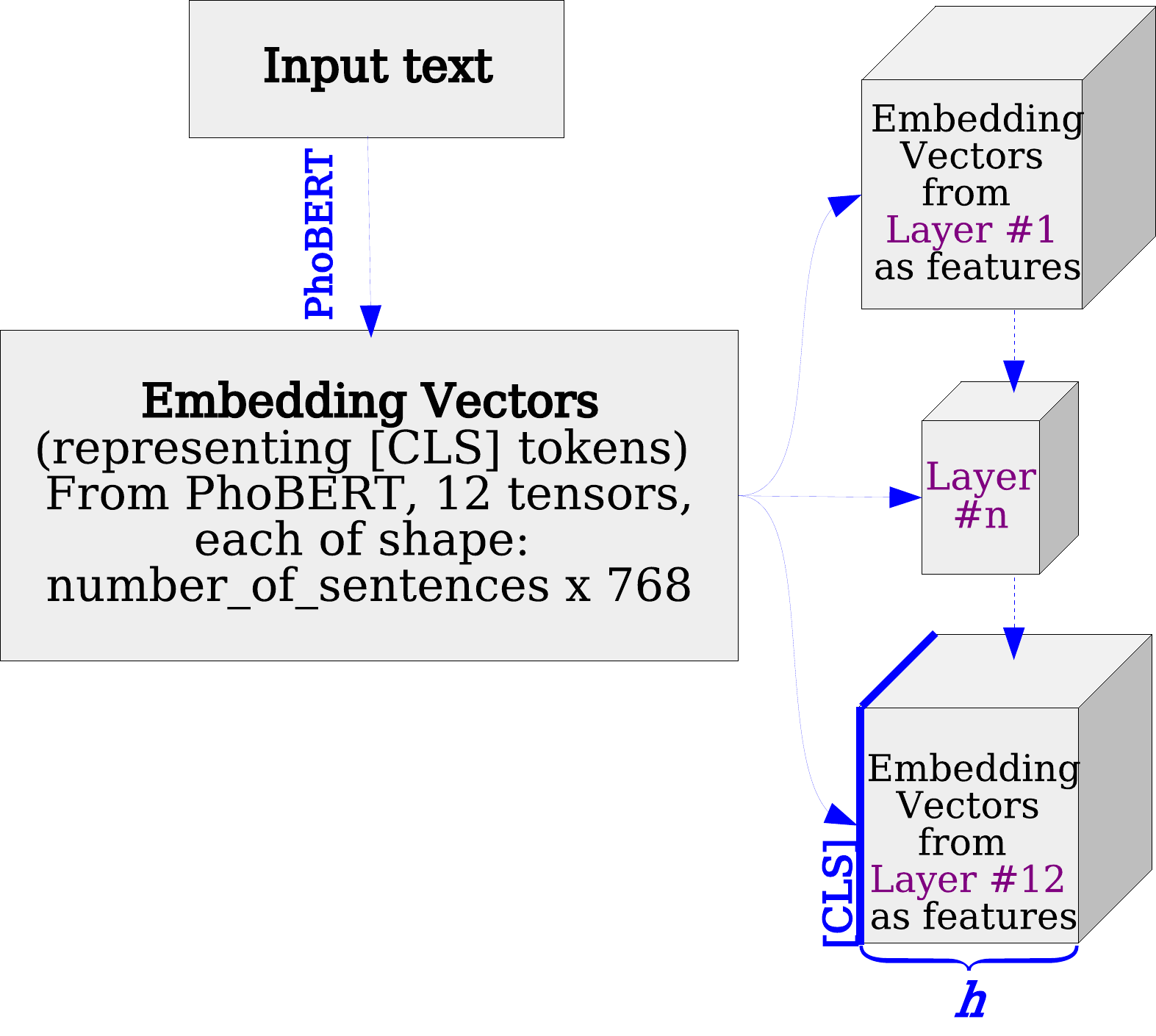}
}\caption{The first token \texttt{[CLS]} as feature of the input sentence} \label{fig:first_CLS}
\end{figure}
In order to fine-tune PhoBERT for downstream tasks, we can use library packages such as Transformers of Hugging Face \citep{wolf-etal-2020-transformers} and FAIRSeq of Facebook \citep{ott2019fairseq} to implement. We use PhoBERT$_{base}$ with 12 block sub-layers of the Encode part to obtain the Embedding vectors as features of input sequences. More specifically, this Embedding vector is an output vector of the first token \texttt{[CLS]} from the final hidden state $\textbf{\textit{h}}$ (Fig.\ref{fig:first_CLS}). According to the idea of \citep{sun2019finetune,neeraj2020bertlayers}, the vector of token \texttt{[CLS]} is the feature of the whole sentence for Classification task. To verify the idea just mentioned earlier, let's consider the following three sentences in Table \ref{table:embeddingCLS}. 
\begin{table}[htb]
\caption{\centering Three Vietnamese sentences are used as examples to extract Embedding vectors with PhoBERT model. Embedding vectors of sentences \textbf{A},\textbf{B} and \textbf{C} are E$_\texttt{[CLS]} ^ \text{A}$, E$_\texttt{[CLS]} ^ \text{B}$ and E$_\texttt{[CLS]} ^ \text{C}$, respectively}  
\centering 
\begin{tabular}{c c c c c} 
\hline 
\parbox{1.5cm}{Sentence}&\parbox{2cm}{\centering Embedding Vector} & Vietnamese & English \\ [0.5ex] 
\hline 
\textbf{A}&\parbox{2cm}{\centering E$_\texttt{[CLS]} ^ \text{A}$} & \parbox{5.5cm}{\vn hà\_nội là thủ\_đô của việt\_nam} & \parbox{5.5cm}{hanoi is the capital of vietnam}\\ \hline
\textbf{B}& \parbox{2cm}{\centering E$_\texttt{[CLS]} ^ \text{B}$} & \parbox{5.5cm}{\vn thủ\_đô của nước chxhcn việt\_nam có tên gọi là hà\_nội} & \parbox{5.5cm}{the capital of the socialist republic
of vietnam is called hanoi}\\
\hline
\textbf{C}& \parbox{2cm}{\centering E$_\texttt{[CLS]} ^ \text{C}$} & \parbox{5.5cm}{\vn hôm\_nay trời sẽ có mưa dông, gió mạnh} & \parbox{5.5cm}{today there will be thunderstorms and strong winds}\\[1ex] 
\hline 
\end{tabular}
\label{table:embeddingCLS} 
\end{table}
\begin{statement}
Assume that Embedding vectors $\text{\textbf{E}}_\texttt{[CLS]} ^ i$ and $\text{\textbf{E}}_\texttt{[CLS]} ^ j$ represent the whole sentences $i$ and $j$, respectively. If sentences $i$ and $j$ are similar, then the Cosine Similarity between $\text{\textbf{E}}_\texttt{[CLS]} ^ i$ and $\text{\textbf{E}}_\texttt{[CLS]} ^ j$ must be sufficiently larger than a certain threshold and converges to 1 with identical sentences.
\end{statement}
\begin{proof}
The Cosine Similarity Formula is
\begin{equation}
Cosine\_Similarity \left( \text{\textbf{E}}_\texttt{[CLS]} ^ i,\text{\textbf{E}}_\texttt{[CLS]} ^ j \right)=\frac{\text{\textbf{E}}_\texttt{[CLS]} ^ i \cdot \text{\textbf{E}}_\texttt{[CLS]} ^ j}{\parallel \text{\textbf{E}}_\texttt{[CLS]} ^ i  \parallel \parallel \text{\textbf{E}}_\texttt{[CLS]} ^ j \parallel}.
\end{equation} 
The statement above can be easily demonstrated through the example in Table \ref{table:embeddingCLS}. As observed, sentences \textbf{A}, \textbf{B} are almost similar and have higher similarity over sentence \textbf{C}. The computation of cosine similarity between Embedding vectors E$_\texttt{[CLS]} ^ \text{A}$ , E$_\texttt{[CLS]} ^ \text{B}$ and E$_\texttt{[CLS]} ^ \text{C}$ is shown in Table \ref{table:value_embeddingCLS}. Since \textbf{A}, \textbf{B} are almost alike, their similarity metric will be high and converge to \textit{1} and vice versa for \textbf{C}.
\end{proof}
\begin{table}[htb]
\caption{\centering Compute Cosine Similarity between Embedding vectors for Table \ref{table:embeddingCLS} with PhoBERT} 
\centering 
\begin{tabular}{c c c c} 
\hline 
\multicolumn{4}{c}{Cosine Similarity(E$_\texttt{[CLS]} ^ i $, E$_\texttt{[CLS]} ^ j $)} \\ [0.5ex] 
\hline 
E$_\texttt{[CLS]} ^ \text{A}$, E$_\texttt{[CLS]} ^ \text{A}$& 
E$_\texttt{[CLS]} ^ \text{A}$, E$_\texttt{[CLS]} ^ \text{B}$& 
E$_\texttt{[CLS]} ^ \text{A}$, E$_\texttt{[CLS]} ^ \text{C}$& 
E$_\texttt{[CLS]} ^ \text{B}$, E$_\texttt{[CLS]} ^ \text{C}$\\
\hline
1.0&0.8519334& 0.4461445& 0.43029878 \\ [0.5ex]
\hline
\end{tabular}
\label{table:value_embeddingCLS} 
\end{table}

\begin{figure}[!htb]
\centering \resizebox{0.5\textwidth}{!}{
\includegraphics[]{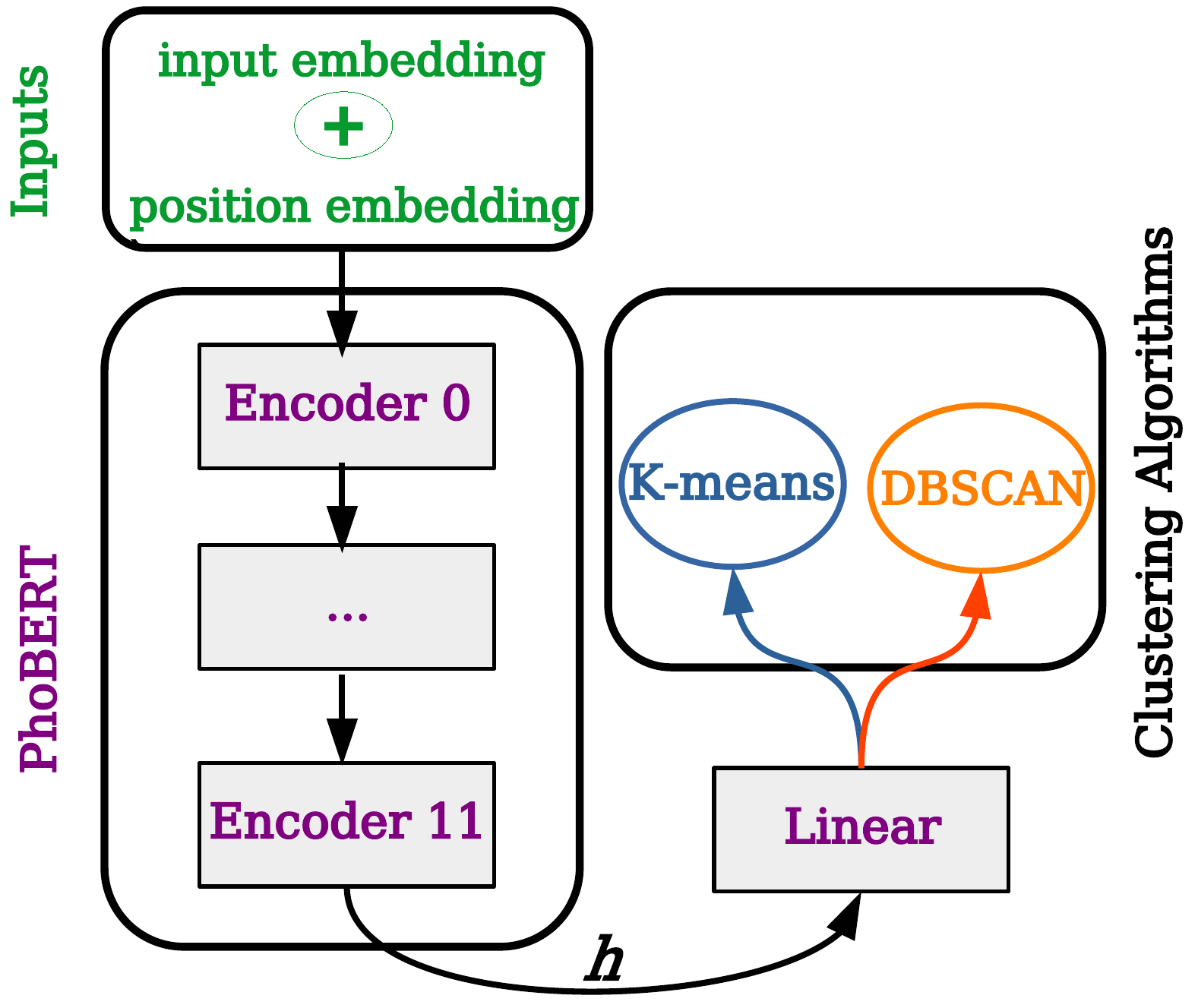}
}
\caption{The overall flow for Clustering task with PhoBERT$_{base}$ model. Starting from Vietnamese pre-training data, passing the layers Encoder$_{0 \rightarrow 11}$ to obtain embedding vectors from the final hidden state $h$ through the Linear projection layer. Finally using unsupervised Learning algorithm K-mean and DBSCAN to cluster the text data.
}\label{flow_clustering}
\end{figure}
Using the Embedding vector of token \texttt{[CLS]} as a feature of the whole sentence, we adapt this idea to our Clustering task. The Clustering implementation process with PhoBERT$_{base}$ model is shown in Figure \ref{flow_clustering}. After obtaining the output embeddings of sentences with PhoBERT$_{base}$ model, we use the algorithm K-mean and DBSCAN to cluster our text data. The output embeddings of sentences have the form as follows
\begin{equation}
\textbf{E}_\texttt{[CLS]} ^ i=\textbf{h}\textbf{W}
\end{equation}
where $\textbf{W} \in \Re ^{d,H}$ and $\textbf{h}$ are projection matrices at the linear projection layer and the final hidden state, respectively.

\section{Clustering algorithm}\label{sec:4}
K-Means is one of the most basic algorithms in unsupervised learning \citep{scikit-learn}. According to the K-Means algorithm, a set of  $N$ samples $\textbf{E}_\texttt{[CLS]}$ is divided into $K$ disjoint clusters ($K<N$). Let $Y$ is the set of all label vectors for  $N$ samples, i.e., each sample $\textbf{E}_\texttt{[CLS]} ^ i$ has a label vector $y_i=[y_{i1}, y_{i2},\ldots,y_{iK}]\in Y$. If vector $\textbf{E}_\texttt{[CLS]} ^ i$ belongs to cluster $k$, then $y_{ik}=1$ and $y_{ij}=0, \forall i \neq k$. Each cluster is characterized by a cluster ``centroids''. A set of centroids is denoted $M=[\textbf{m}_1,\textbf{m}_2,\ldots, \textbf{m}_K]$. In K-means algorithm, the clustering problem will be reduced to the optimization for loss function $\mathcal{L}\left(\textbf{Y}, \textbf{M} \right)$.
\begin{equation}
\mathcal{L}\left(\textbf{Y}, \textbf{M} \right)= \sum \limits _{i=1}^{N} \sum \limits _{j=1}^{K} y_{ij} \left|\left| \textbf{E}_\texttt{[CLS]}^i -\textbf{m}_j \right|\right|^2
\end{equation}

\RestyleAlgo{ruled}\begin{algorithm}[htb]
\SetKwInOut{Input}{Input}\SetKwInOut{Output}{output}
\caption{Pseudocode of original DBSCAN algorithm for our data}\label{alg:one}
\KwData{$\textbf{E}_\texttt{[CLS]}$}
\Input{$\varepsilon$, $\textit{MinPts}$}
\Input{$metric$\tcp*[f]{calculating distance between data points}} 
\Input{\textbf{cluster}}
\ForEach(\tcp*[f]{linear scan for all data points in $\textbf{E}_\texttt{[CLS]}$}){$p$ in $\textbf{E}_\texttt{[CLS]}$}{
	\lIf (\tcp*[h]{\textit{p} is unassigned to any \textit{cluster} or \textit{noise}}) {cluster($p$) $!=$ unassigned} {\textbf{continue}}
	Neighbors $N \leftarrow$ find\_neighbors($\textbf{E}_\texttt{[CLS]}$, $metric$, $p$, $\varepsilon$)\\
	\If() {$|N| < minPts$}{
		\textbf{cluster}($p$) $\leftarrow$ \textit{noise}\\
		\textbf{continue}
	}
	$c \leftarrow $ create a new cluster\\
	\textbf{cluster}($p$) $\leftarrow$ $c$\\
	\textbf{\textit{S}} $\leftarrow$ $N \setminus \{p\}$\\
	\ForEach {$q$ in \textbf{S}}{
		\lIf{\textbf{cluster}($q$)$==$\textit{noise}}{\textbf{cluster}($q$) $\leftarrow$ $c$}
		\lIf {cluster($q$) $!=$ unassigned} {\textbf{continue}}
		Neighbors $N \leftarrow$ find\_neighbors($\textbf{E}_\texttt{[CLS]}$, $metric$, $q$, $\varepsilon$)\\
		\textbf{cluster}($q$) $\leftarrow$ $c$\\
		\lIf() {$|N| < minPts$}{\textbf{continue}}
		\textbf{\textit{S}} $\leftarrow$ $N \cup \textbf{\textit{S}}$		
	}
}
\end{algorithm}
Along with that, we also apply DBSCAN technique to cluster data points \citep{ester1996densitybased}. In real-life data, DBSCAN can work well for nonconvex clusters with arbitrary shapes and noises. DBSCAN  algorithm focuses on radius \textit{eps}--$\varepsilon$ and the minimum number of neighbors required to create a cluster $minPts$. Radius \textit{eps} defines a circle for each point to determine its neighbors. A point becomes \textit{core point} if the circle surrounding this point with radius \textit{eps} contains more than \textit{MinPts} neighbors. In case the number of neighbor points is less than \textit{MinPts}, the \textit{core point} is the \textit{border points}. On the other hand, a point without any neighbors within radius \textit{eps} is called \textit{noise}. The relationship state of two points in DBSCAN can be \textit{direct density reachable}, \textit{density reachable} or \textit{density connected}. A point is called \textit{direct density reachable} for $C_i$ point if and only if it lies within the circle centered \textit{core point} $C_i$. If a \textit{core point} is connected unidirectionally to any other \textit{core point} through a chain of \textit{core points}, there is a \textit{density reachable} state between them. In case there are two points, which are \textit{density reachable} from the same point, they are \textit{density connected} states. Pseudocode describing DBSCAN clustering algorithm \citep{DBLP:journals/tods/SchubertSEKX17,ester1996densitybased} is shown in Algorithm \ref{alg:one}.

Unlike the evaluation metrics for supervised learning algorithms, the evaluation of clustering performance can be applied to datasets with known or unknown ground truth labels. If the ground truth class assignment of dataset is known, we use entropy-based measure, \textbf{V-measure} proposed in \citep{rosenberg-hirschberg-2007-v} to evaluate clustering performance for our sample dataset. The sample dataset will be described in detail in Section \ref{sec:exp}. Based on the conditional entropy analysis of two terms of \textit{homogeneity} and \textit{completeness}, V-measure is a harmonic mean function of those terms and can be calculated as follows 
\begin{equation}
v(\beta, h, c)=\frac{(1+\beta)\times h \times c}{(\beta \times h)+c}, \label{eq:v-measure}
\end{equation}
where $h=\frac{1-H(C|K)}{H(C)}$ and $c=\frac{1-H(K|C)}{H(K)}$ are \textit{homogeneity} and \textit{completeness} respectively. The conditional entropy $H(K|C)$ and entropy $H(K)$ are symmetric. In formula \eqref{eq:v-measure}, $\beta$ weight represents the contributions of homogeneity or completeness and the default value of $\beta$ is equal to 1. 

Unfortunately, in fact, we don't know anything about the ground truth classes for document clustering task. Thus, we can evaluate clustering performance based on the partition obtained from clustering techniques and two types of proximities, which are similarity or dissimilarity between objects. Suggested in \citep{ROUSSEEUW198753} \textbf{Silhouette} is a typical evaluation for this case. Besides providing a graphical overview of the partitional clustering (silhouette plot), Silhouette also allows evaluating clustering validity based on the average silhouette width. From those analyses, we can obtain a suitable number of clusters for the K-means algorithm. The Silhouette Coefficient $s(i)$ for object $i$ has the form
\begin{equation}
s(i)=\frac{b(i)-a(i)}{max \{ a(i), b(i) \} }.
\end{equation}
Here $a(i)$ is the average dissimilarity of object $i$ to the remaining objects in the same cluster and $b(i)$ is the average dissimilarity of $i$ to all objects of the next nearest cluster. The value of Silhouette Coefficient is in the range $[-1,+1]$, where near -1 indicates that the object for incorrect clustering and vice versa for +1. The value around 0 represents overlapping clusters.
\section{Experiment} \label{sec:exp}
We apply K-means and DBSCAN algorithm with PhoBERT$_{base}$ to cluster our Facebook Vietnamese conversations dataset (FVnC) and Sample dataset. In order to implement clustering task, we use  PhoBERT$_{base}$ with the Transformers package of Hugging Face and Scikit-learn library \citep{wolf-etal-2020-transformers, scikit-learn-clustering}. Furthermore, we also search the optimal parameters for the clustering algorithms in this article. The clustering results will be used to build Intents for chatbot later.
\subsection*{Clustering task dataset}
\begin{figure}[!htb]
\centering \resizebox{0.6\textwidth}{!}{
\includegraphics[]{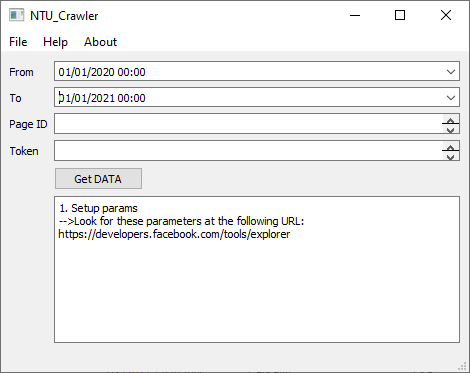}
}\caption{User interface of \textit{NTUCrawler} on Windows}\label{fig:UI_NTUCrawler}
\end{figure}

We evaluate our approach on Facebook Vietnamese conversations dataset. There are plenty of free tools or extensions to download conversations from a personal page because it is quite simple. However collecting conversations from public page is more difficult so most tools or extensions to carry out this task are paid. 
In order to acquire this dataset, we created a tool named \textit{NTUCrawler}\footnote{Tool is available at \url{https://archive.org/download/NTUCrawler}} for scraping conversations from a Facebook messenger page of our University. This tool is written in Python language and based on Facebook's Graph API platform to get messages. It has two versions, one is linux executable (run on Ubuntu distribution) and the other is Windows executable. The UI of the Windows version is shown in Figure \ref{fig:UI_NTUCrawler}. \textit{NTUCrawler} requires users to provide four parameters, start time and finish time to get data, PageID and Token of page. Downloaded dataset contains 8000 conversations with more than 150 thousand raw text sentences of clients and admins of a Facebook page in the six-month period of the year 2020. The contents of the conversations are FAQ (frequently asked questions), which are related to information already published on the university website, for example, tuition fees, insurance, dormitory, English language test, course registration, etc.

\subsection*{Data preprocessing}
The raw downloaded dataset above must be preprocessed. Some punctuations ``\texttt{!}'', ``\texttt{"}'', ``\texttt{?}'', ``\texttt{'}'',  ``\texttt{:}'' are removed from the dataset. Further, we eliminate stop words that have no value or no meaning to the NLP model such as ``\texttt{dạ}'', ``\texttt{vâng}'', ``\texttt{chào ad}'', ``\texttt{vâng ạ}'', ``\texttt{alo}'', ``\texttt{\vn ừ}'',  ``\texttt{\vn vậy}'', ``\texttt{\vn ok}'', ``\texttt{\vn nhé}''. We also filter out duplicate sentences or those that contain less than 4 words. After data preprocessing, 44846 text sentences are obtained and it is our desired \textbf{FVnC} dataset. From FVnC dataset, we randomly selected 95 sample text sentences (0.2\% of FVnC dataset size) to form the \textbf{Sample} dataset. Besides that, we also use another sub dataset which is called the \textbf{wiki} dataset. This dataset contains 396 text sentences of articles on 5 topics collected from Wikipedia. The Wikipedia library was applied to access and parse data from Wikipedia. The reason we use those Sample dataset and wiki sub dataset is that we can evaluate the effectiveness of applying PhoBERT$_{base}$ for downstream task (clustering). Label assignment to sample text sentences and   analyzing the number of clusters will cost less in our task. In addition, clustering on the small Sample dataset not only helps to easily evaluate clustering performance but also saves time compared to the original FVnC dataset. These sample text sentences were completed manual labeling by us and divided into 3 classes, which describe questions between users (students) and admin (university) about information related to insurance, dormitory and English language test. Specifically, in order to specify the lables of classes, we rely on the experience of the specialists of the training department, who are responsible for answering students' questions directly or via social platforms. In their opinion, first-year students of our university who are often concerned with insurance, dormitory and English language test. During data collection for each label, we carefully selected the sentences in the dataset that matched the recommendations of the specialists. Those 3 labels are one of the intents used to train the chatbot. Table \ref{table:dataset} shows the details of the two datasets. 
\begin{table}[htb]
\caption{\centering Brief description of the datasets used for PhoBERT$_{base}$} \label{table:dataset} 
\centering
\resizebox{\textwidth}{!}{
\begin{tabular}{c c c c c}
\hline 
Dataset & Label & Description & Tasks & $\textbf{E}_\texttt{[CLS]}^i$ size \\ 
\hline 
FVnC & unknown & \parbox[][1cm][c]{5cm}{\centering clean downloaded dataset} & \parbox{4cm}{\centering Silhouette evaluation, clustering} & 44846 $\times$ 768\\ 
\hline 
      & class 1 & \parbox[][1.2cm][c]{5cm}{\centering features relating to \textit{insurance}} &  & 31 $\times$ 768\\ 
Sample& class 2 & \parbox[][1.2cm][c]{5cm}{\centering features relating to \textit{english language test}} & \parbox[][1cm][c]{4cm}{\centering clustering, clustering performance evaluation}& 38 $\times$ 768\\ 
      & class 3 & \parbox[][1.2cm][c]{5cm}{\centering features relating to \textit{dormitory}} &  & 26 $\times$ 768\\ 
\hline 
\end{tabular} 
}
\end{table}

\noindent According to \citep{DBLP:journals/corr/abs-2003-00744}, input text must be already word-segmented before going through the BPE algorithm. We use ``\texttt{pyvi}'' toolkit of Tran \citep{pyvi} to perform word segmentation in our datasets. After passing the fastBPE step, we have the index of tokens and attention masks for the text data. Taking them through PhoBERT's architecture leads to output embeddings of $\textbf{E}_\texttt{[CLS]}$. From this step, we use these embeddings as feature vectors to cluster text data. 
\subsection*{Dealing with varying length}
Because the length of the sentences in FVnC dataset is different, we have to use padding to make sure the input texts have the same length.
\begin{figure}[!htb]
\centering \resizebox{0.5\textwidth}{!}{
\includegraphics[]{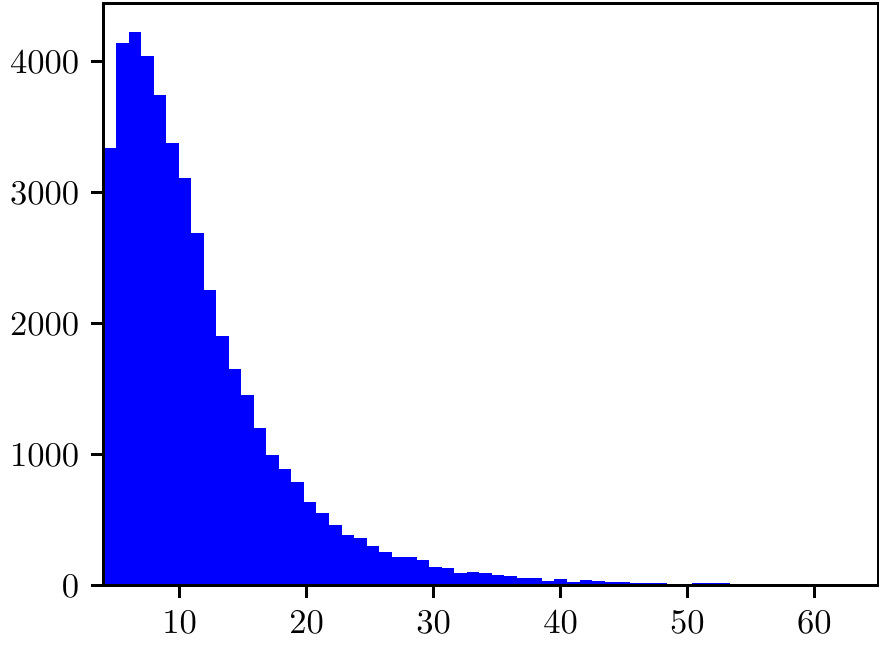}
}\caption{The distribution of the sentence lengths in FVnC dataset}
\label{fig:len_of_sentence}
\end{figure} 
In particular, the maximum sequence length of PhoBERT is 256. We truncate sentences with padding length less than 256 tokens. To choose the optimal padding length for all sentences, we can analyze the distribution of sentence lengths in Figure \ref{fig:len_of_sentence}. Based on the above distribution, most sentences have lengths of less than 33 words. Thus, we decide that the padding length is equal to 33 in all datasets.
\subsection*{Parameter optimization}
When using the DBSCAN technique, we need to take care of two parameters \textit{MinPts} and $\varepsilon$. Choosing these two parameters is not easy. Their influence is very large on the clustering results. There is no way to accurately determine the parameter \textit{MinPts}. However, there are several ways of choosing \textit{MinPts} which have been proposed in \citep{ester1996densitybased,Sander2004DensityBasedCI}. Besides, the value of \textit{MinPts} must also depend on domain knowledge and the data distribution observation. So we derive that \textit{MinPts} should be greater than the number of dimensionality of feature vector $\textbf{E}_{\texttt{[CLS]}} \left( 44846 \times 768 \right)$. As we can see, the number of dimensions of $\textbf{E}_{\texttt{[CLS]}}$ is too large (\textit{768}), which affects computation time and cost for large data sets. Therefore, we use the dimensionality reduction method to reduce $\textbf{E}_{\texttt{[CLS]}}$ to lower-dimensional while it retains most of the original information. Principal Component Analysis (\textbf{PCA}) and t-distributed Stochastic Neighbour Embedding (\textbf{t-SNE}) are common techniques for data dimensionality reduction. \textbf{PCA} relies on eigenvalues and eigenvectors of $\textbf{E}_{\texttt{[CLS]}}$  to reduce the original data to a specific number of dimensions (commonly known as \textit{principal components}) but it still ensures a threshold of allowable variance. If we use \textbf{PCA} then \textit{MinPts} can be selected as follows
\[MinPts \geq \textit{principal components}+1\]
\begin{figure}[!htb]
\centering \resizebox{\textwidth}{!}{
\includegraphics[]{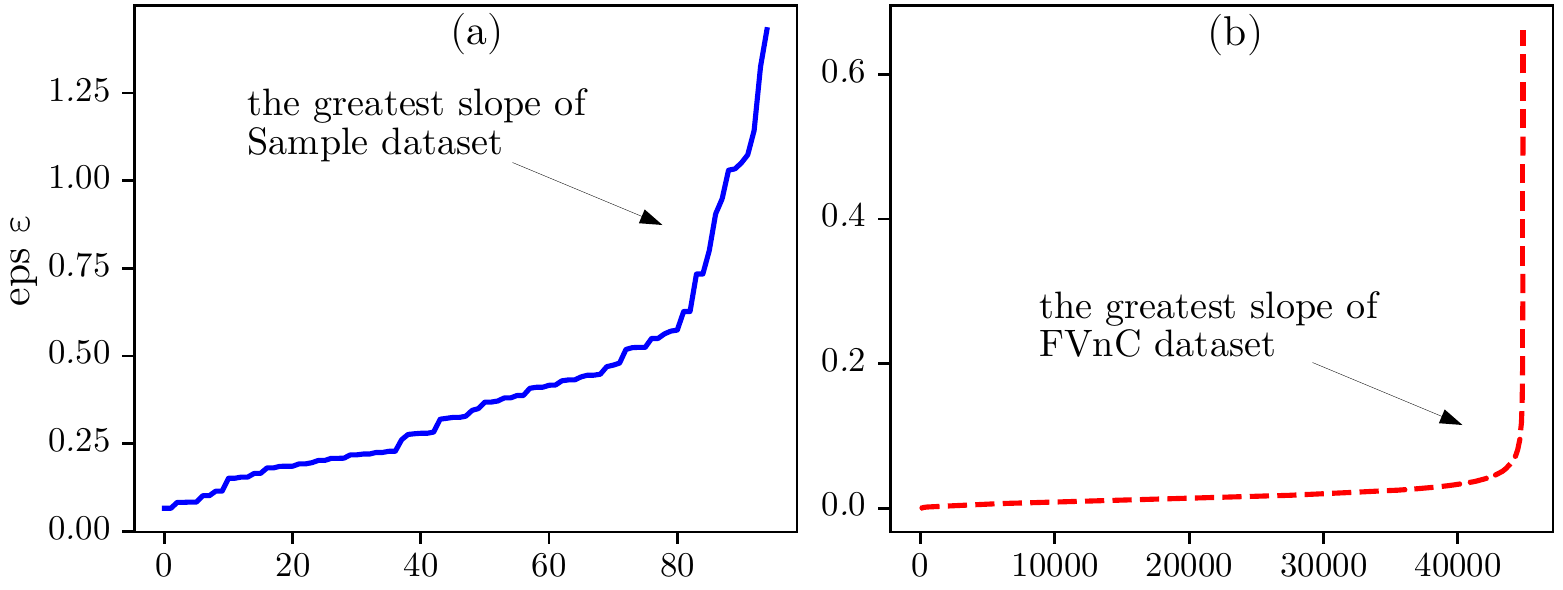}
}\caption{Data Points are sorted ascending by the average distance to \textit{MinPts} nearest neighbors. (a) Calculating on Sample dataset; (b) Calculating on FVnC dataset}\label{fig:kdistance}
\end{figure}
Parameter $\varepsilon$ can be found from a K-Distance graph, which is based on the average distance between objects and its \textit{MinPts} nearest neighbors \citep{Gaonkar2013AutoEpsDBSCAND}. The K-Distance graph with \textit{MinPts=3} for FVnC and Sample dataset is shown in Figure \ref{fig:kdistance}. The blue solid curve and red dashed curve correspond to the average distance of objects to \textit{MinPts} nearest neighbors which are sorted in ascending order for FVnC and Sample dataset respectively. Usually, a point at the position with the largest slope change in K-Distance graph or what we popularly call the ``knee/elbow'' of the graph is the optimal value of parameter $\varepsilon$ \citep{Rahmah_2016}. Especially, the greatest slope change zones are highlighted in Fig.\ref{fig:kdistance}(a) and Fig.\ref{fig:kdistance}(b) for specific datasets. In order to take exactly the point mentioned above or the ``knee point'' of the graph, the kneedle algorithm is considered in our work \citep{satopaa2011finding}. The knee point obtained from the kneedle algorithm is determined by the intersection of the specific data curve with the vertical straight line in Figure \ref{fig:knee_point}. The optimum values for parameter $\varepsilon$ are 0.57 and 0.12 in the case of Fig.\ref{fig:knee_point}(a) and Fig.\ref{fig:knee_point}(b), respectively.
\begin{figure}[!htb]
\centering \resizebox{\textwidth}{!}{
\includegraphics[]{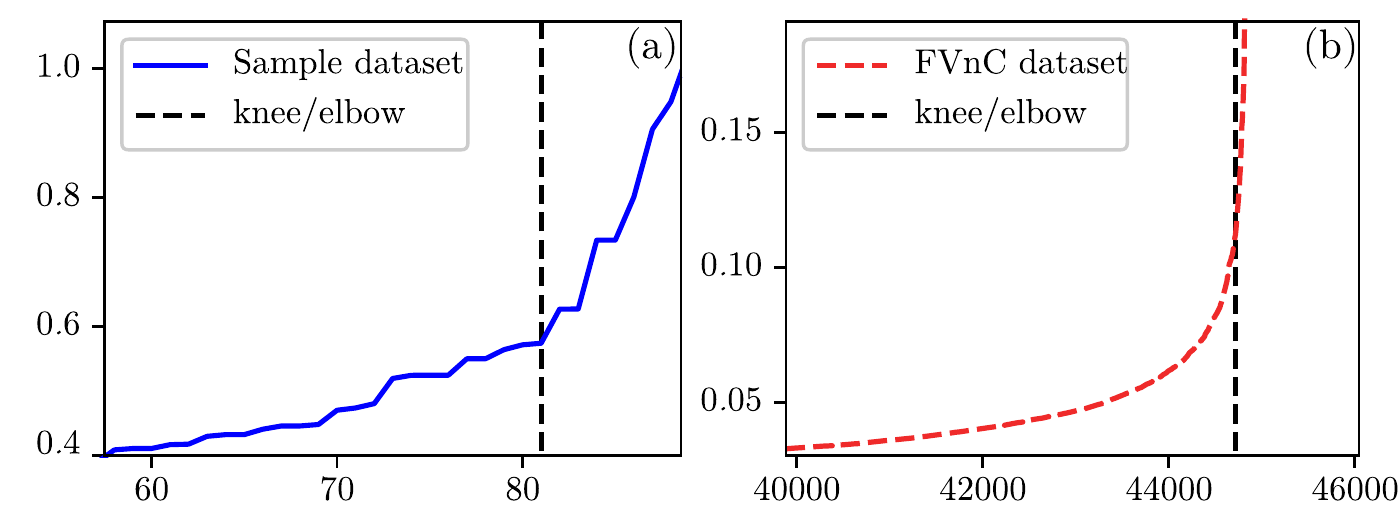}
}\caption{Determining knee points at the greatest slope change zones using the kneedle algorithm for Figure \ref{fig:kdistance}. (a) Solid blue curves -- the greatest slope change zone of Sample dataset; (b) Dashed red curve -- the greatest slope change zone of FVnC dataset.}\label{fig:knee_point}
\end{figure}
\RestyleAlgo{ruled}\begin{algorithm}[htb]
\SetKwInOut{Input}{Input}\SetKwInOut{Output}{Output}
\SetKwInOut{Initial}{Initial}
\SetKwInOut{Data}{Data}
\caption{Pseudocode of the proposed technique to find appropriate $\varepsilon$ and \textit{MinPts}}\label{alg:two}
\Data{$\textbf{E}_\texttt{[CLS]}$, label \tcp*[l]{label: points labels (unassigned or assigned)}}
\Input{K-Distance}
\Input{$n\_components$, step \tcp*[l]{step: minimum distance between points}}
\Output{Index \tcp*[h]{the appropriate $\varepsilon$ and $MinPts$ values for DBSCAN}}
\Initial{$\text{Max}_{Silhouette} \leftarrow -1$ ; $\text{Max}_{Vmeasure} \leftarrow 0$ ; $\text{Max} \leftarrow -0.5$}
\textit{slope} $\leftarrow$ \texttt{CalculatingSlope}(K-Distance)\tcp*[h]{the greatest slope change zone}\\
\textit{nearest\_neighbors} $\leftarrow$ \texttt{arange}($n\_components+1$, $2 \times n\_components+1$, step)\\
\ForEach{$\varepsilon$ in slope}{
	\ForEach{\textit{MinPts} in nearest\_neighbors}{
		$p \leftarrow \texttt{PCA}\left(\textbf{E}_\texttt{[CLS]}, n\_components\right)$\\
		$ClusterAssignment \leftarrow \texttt{DBSCAN} \left(p, \textit{MinPts}, \varepsilon \right)$\\
		$\text{SilCoeff} \leftarrow \texttt{SilhouetteScore}(p, ClusterAssignment)$\\
		\uIf{label $=$ unassigned}{
			\uIf{$\text{SilCoeff} > \text{Max}_{Silhouette}$}{
				$\text{Max}_{Silhouette} \leftarrow \text{SilCoeff}$\\
				$\text{Index}\leftarrow (\varepsilon, MinPts)$			
			}

		}\Else{
			$\text{VScore} \leftarrow \texttt{VMeasureScore} \left(label, ClusterAssignment \right)$\\
			\uIf{$\left(\text{SilCoeff}+\text{VScore} \right)/2>\text{Max}$}{
				$\text{Max} \leftarrow \left(\text{SilCoeff}+\text{VScore} \right)/2$\\
				$\text{Index}\leftarrow (\varepsilon, MinPts)$
			}
		}					
	}		
}
\end{algorithm}
However, optimizing parameter $\varepsilon$ by choosing a fixed value of knee point in some cases does not lead to good clustering efficiency. Specifically in the test of the good separability between clusters in the subsection below, Silhouette score is quite low. This could lead to objects being assigned to the wrong clusters. With the expectation of improved clustering performance, in this work, we propose another technique based on the combination of K-Distance graph and clustering performance evaluations to find the right optimal value $\varepsilon$. Besides observing the greatest slope change zone of the line from a pair of points on the K-Distance graph, the maximum value of the Silhouette coefficient and V-measure score are also considered. The pseudocode for our technique is given in Algorithm \ref{alg:two}. The sequence of \textit{MinPts} is taken from $\textit{principal components}+1$ to $2\times \textit{principal components}+1$ and incremented by a step of the minimum distance between data points. Gridsearch technique is applied in algorithm \ref{alg:two} for the sequence of \textit{MinPts} and the greatest slope change zone in K-Distance graph. At the position where the Silhouette Coefficient is maximum, we obtain the optimal pair of values $\left(\varepsilon, MinPts \right)$ for unlabeled data. For the labeled Sample dataset, the search of the optimal values $\left(\varepsilon, MinPts \right)$ is based on the greatest mean of V-measure and Silhouette evaluation.

\subsection*{Result}
\begin{figure}[!htb]
\centering \resizebox{0.8\textwidth}{!}{
\includegraphics[]{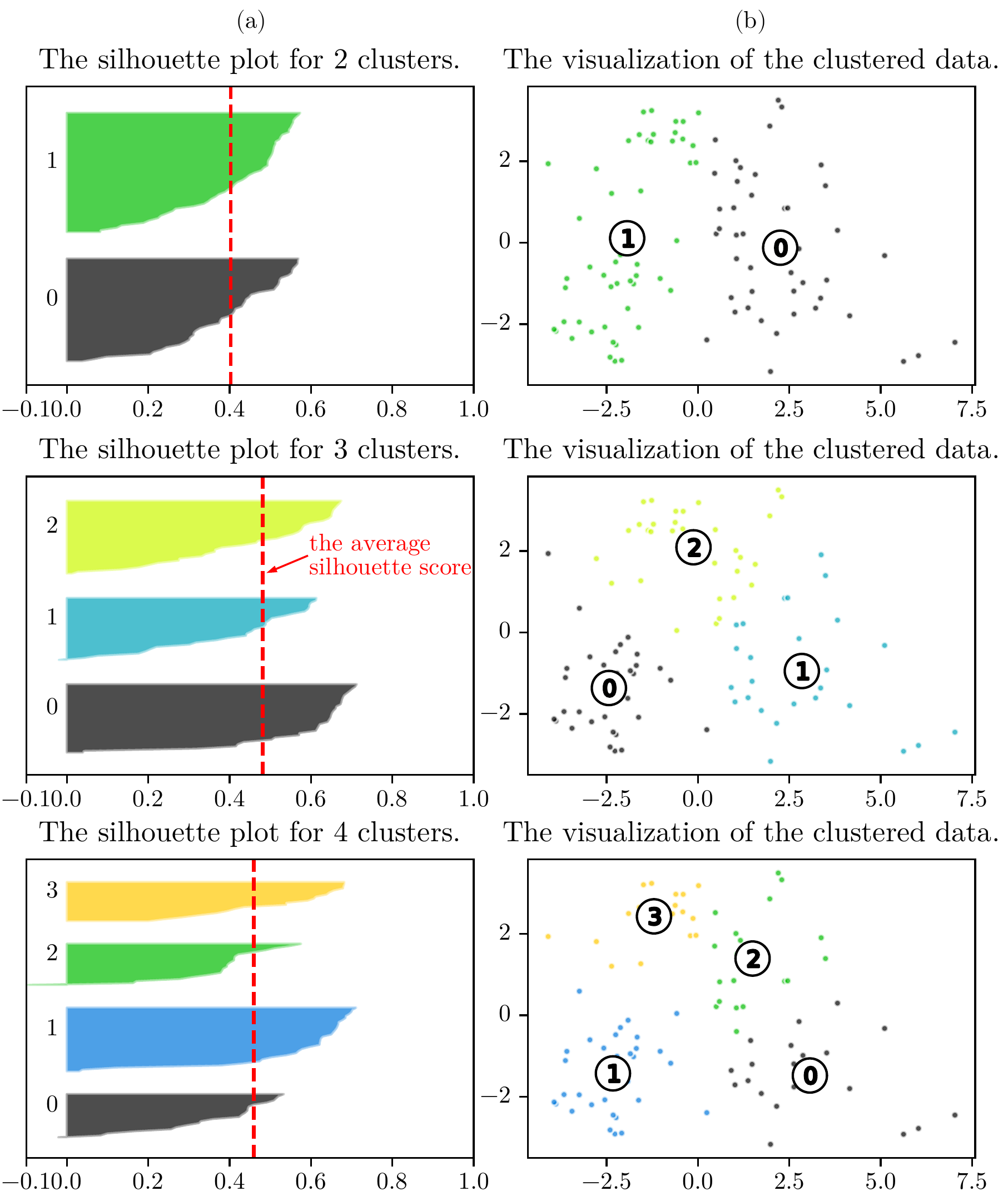}
}\caption{The graphical overview of the partitional clustering for 2, 3, 4 clusters using K-Means algorithm on the Sample dataset. (a) The silhouette plot; (b) The visualization of the clustered data for corresponding silhouette plots.}\label{fig:silhouette_sample_data}
\end{figure}
In this part, our first task is to cluster text documents and evaluate clustering performance on the Sample dataset. As a consequence, analyzing on the Sample dataset will be generalized to the general dataset FVnC such as choosing the number of clusters using silhouette analysis for K-Means algorithm. As mentioned above, the Sample dataset has three labeled clusters (see Table \ref{table:dataset}). We use Silhouette evaluation to confirm that the Sample dataset has exactly three clusters and the way to choose the right number of clusters when using it. Besides considering average silhouette scores, the silhouette plot is also an important factor in determining the number of clusters. Figure \ref{fig:silhouette_sample_data} represents the graphical overview of the partitional clustering using K-Means algorithm on Sample dataset. X-axis of Fig.\ref{fig:silhouette_sample_data}(a) corresponds to the silhouette coefficient values and the vertical dashed line is the average of the silhouette coefficients of data points. Clusters corresponding to silhouette plots in Fig.\ref{fig:silhouette_sample_data}(b) are visualized in two dimension space using PCA. The maximum value of average silhouette scores is close to 0.481 for 3 clusters. On the other hand, the thickness of the silhouette plot for clusters is similar. The analysis outlined above fits the facts in Table \ref{table:dataset} for the Sample dataset.


\begin{table}[!htb]
\caption{\centering Evaluating feature extraction efficiency of models through V-measure scores for clustering task on the Sample dataset and wiki dataset.} \label{table:comparison_models} 
\centering \begin{tabular}{l c c}
\hline
\multirow{2}{*}{Approach} & \multicolumn{2}{c}{Dataset} \\ \cline{2-3} &  Sample & wiki\\
\hline
FastText (cc.vi.300) & 0.64 & 0.27\\
GloVe (glove.6B.100d) & 0.49 & 0.26\\
GloVe (vncorpus.3B.100d) & 0.61 & 0.27\\
BERT$_{base}$ Uncased & 0.11 & 0.19\\
BERT$_{base}$ Multilingual Uncased & 0.36 & 0.43\\
DistilBERT$_{base}$ Multilingual Cased & 0.37 & 0.12\\
GPT-2 & 0.04 & 0.04\\
PhoBERT$_{base}$ & \textbf{0.76} & \textbf{0.62}\\
\hline 
\end{tabular} 
\end{table}

Moreover, in order to evaluate the efficiency of the PhoBERT$_{base}$ model in feature extraction for clustering tasks, we compared it with the other models such as BERT$_{base}$ Uncased, BERT$_{base}$ Multilingual Uncased \citep{DBLP:journals/corr/abs-1810-04805}, DistilBERT$_{base}$ Multilingual Cased \citep{Sanh2019DistilBERTAD}, GPT-2 \citep{radford2019language} based on V-measure score, among which GPT-2 and BERT$_{base}$ Uncased models do not support Vietnamese. We also use traditional approaches like FastText, GloVe to compare with transformer models. Both models are applicable to Vietnamese language. However, both of these models are commonly used for learning word representations. Thus, to obtain sentence embedding, our method is averaging the word embeddings of all the words in the sentence. Especially, in order to make the comparison better, we also retrained the GloVe model on a new Vietnamese 20GB dataset based on the improvement proposed in \citep{nguyen2020analyze}.  Pre-trained word vectors \textbf{vncorpus.3B.100d} of new GloVe model corresponds to the corpus 3B tokens, 1.3M vocab, and 100d vectors and 1.17 GB download. Comparison results on the Sample dataset and wiki dataset using K-Means algorithm with 3 clusters and 5 clusters respectively are presented in Table \ref{table:comparison_models}. Through the obtained results, the models that support Vietnamese achieve significantly better V-measure scores than the rest of the models, especially the PhoBERT$_{base}$ model, which obtained the highest V-measure score on both the Sample dataset and wiki dataset (0.76 and 0.62 respectively). The comparisons obtained are in good agreement with the pointed theoretical and experimental works. Therefore, we use PhoBERT$_{base}$ model to extract features for the FVnC dataset, which leads to the output embeddings serving the clustering task.

\begin{table}[!htb]
\caption{\centering Experiment to find the optimal values of parameters $\varepsilon$, \textit{MinPts} for DBSCAN algorithm on the Sample dataset.} \label{table:optimal_value_on_Sample} 
\centering
\resizebox{\textwidth}{!}{
\begin{tabular}{c c c c c c c}
\hline 
Approach & eps ($\varepsilon$) & \textit{MinPts} &  V-measure score & Silhouette score & Average & N-Clusters \\ 
\hline 
\multirow{5}{*}{Algorithm \ref{alg:two}} & \textbf{0.84}	& \textbf{4}	& \textbf{0.56}	&\textbf{0.28}	&\textbf{0.42}	&\textbf{5} \\  
& 0.78&	4&	0.52&	0.27&	0.4&	5 \\ 
&0.72&	4&	0.5&	0.22&	0.36&	6 \\  
&0.76&	3&	0.47&	0.22&	0.34&	4\\
&0.54&	5&	0.37&	0.0&	0.18&	8\\
\hline 
\multirow{3}{*}{Knee point} & 0.57&	3&	0.42&	0.15&	0.29&	9\\ 
&0.57 &4	& 0.41&	0.1&	0.26&	8\\
& 0.57&	5	&0.41	&0.05	&0.23	&7\\
\hline
\end{tabular} 
}
\end{table}


\begin{figure}[!htb]
\centering \resizebox{\textwidth}{!}{
\includegraphics[]{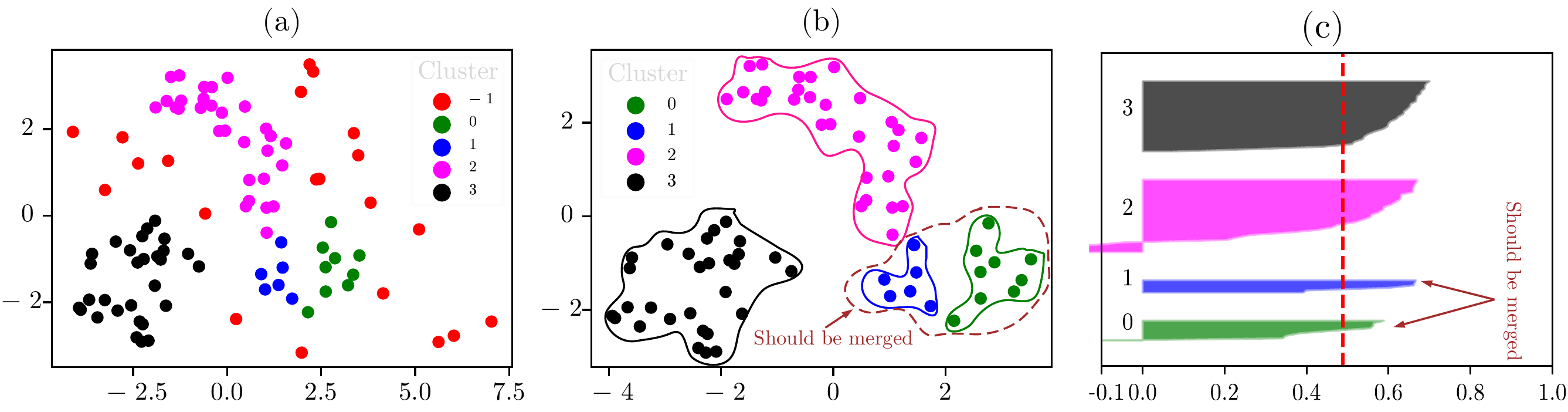}
}\caption{DBSCAN clustering on the Sample dataset}\label{fig:dbscan_sampledata}
\end{figure}
For DBSCAN clustering  method on the Sample dataset, we need to find the optimal parameters $\varepsilon$,  \textit{MinPts} using Algorithm \ref{alg:two} and kneedle algorithm. Based on K-Distance graph of Sample dataset in Fig.\ref{fig:kdistance}(a) and knee visualization in Fig.\ref{fig:knee_point}(a), the greatest slope zone is selected in the range of $[0.5,0.85]$ and the knee point value is 0.57. Due to ``\textit{principal components}'' equal to 2 in PCA dimensionality reduction, the value of \textit{MinPts} will be selected from 3 to 5. Some experimental results in finding the optimal parameters are shown in Table \ref{table:optimal_value_on_Sample}. As shown in the table, $\varepsilon$, \textit{MinPts} and the average of V-measure and Silhouette score equal to 0.84, 4 and 0.42 respectively are the best choice to cluster data points. Results in Table \ref{table:optimal_value_on_Sample} also show that our approach -- Algorithm \ref{alg:two} gives better clustering performance than kneedle algorithm on the Sample dataset. Using DBSCAN algorithm with the obtained parameters, 5 clusters are formed in Figure \ref{fig:dbscan_sampledata}. In which cluster ``-1'' contains noisy objects in Fig.\ref{fig:dbscan_sampledata}(a), noisy objects are removed in Fig.\ref{fig:dbscan_sampledata}(b)  and Silhouette plot of denoised Sample dataset in Fig.\ref{fig:dbscan_sampledata}(c).
After removing noise, our clustering result has 4 clusters while the Sample dataset has only 3 clusters. Based on actual observation in Fig.\ref{fig:dbscan_sampledata}(b) and Silhouette plot in Fig.\ref{fig:dbscan_sampledata}(c), cluster 0 and 1 must be merged into one cluster.   
\vspace*{3mm}
\begin{remark} 
DBSCAN is a density-based spatial clustering algorithm. The biggest disadvantage of DBSCAN is working in cases of varying density clusters. From the Silhouette plot in Fig.\ref{fig:dbscan_sampledata}(c), there are two separate clusters of very different density comparing to the other two clusters. Two low-density clusters (cluster 0 and 1) are a matter of concern to us for predicting the number of clusters. Obviously, we hope that our clusters will be more equally distributed to choose the number of clusters more precisely. In order to solve this problem, we propose some solutions as follows\\
\textbf{(i)} Based on the thickness of the silhouette plot, the silhouette plots for clusters 0 and 1 are much smaller than clusters 2, 3. In particular, clusters 0 and 1 are close neighbors. Therefore, we have reasons to merge these small two clusters into a larger cluster. As a result, 3 clusters are the correct number of clusters when using DBSCAN clustering on the Sample dataset. After all, we should combine the selection of the optimal parameters and the width of Silhouette plots for clusters when using DBSCAN algorithm to obtain the most accurate number of clustering.\\
\textbf{(ii)} On the other hand, we can use a criterion for determining the minimum number of core points in a cluster. If at least two clusters are close to each other and have a smaller the minimum number of core points than the specified criteria, we can merge them into a larger cluster.
\end{remark}
\begin{table}[!htb]
\caption{\centering Silhouette scores using K-Means algorithm on original FVnC and denoised FVnC datasets for several different cluster numbers were found from DBSCAN algorithm.} \label{table:silhouette_scores}
\centering
\resizebox{0.75\textwidth}{!}{
\begin{tabular}{c c c c c}
\hline 
\multicolumn{3}{c}{DBSCAN} &  \multicolumn{2}{c}{\parbox[][1.1cm][c]{5cm}{\centering K-Means\\Silhouette Scores}} \\ 
\hline 
eps $\varepsilon$ & \textit{MinPts} & N-Clusters & Original FVnC  & Denoised FVnC\\ 
\hline 
0.14  &  22  &  10  &  \textbf{0.3359}    &   0.3436 \\ 

0.14  &  25  &  12  &  \textbf{0.3359}    &   \textbf{0.3460} \\ 

0.13  &  12  &  15  &  0.3310    &   0.3368 \\ 

0.14  &  19  &  16  &  0.3294    &   0.3326 \\

0.13  &  16  &  20  &  0.3267    &   0.3314 \\ 

0.12  &  30  &  23  &  0.3252    &   0.3418 \\ 

0.11  &  26  &  29  &  0.3225    &   0.3343 \\ 

0.1  &  14  &  53  &  0.3196    &   0.3257 \\ 

0.09  &  12  &  75  &  0.3196    &   0.3275 \\ 

0.1  &  3  &  81  &  0.3217    &   0.3205 \\ 

0.09  &  4  &  102  &  0.3220    &   0.3229 \\ 

0.09  &  3  &  128  &  0.3207    &   0.3228 \\ 

0.08  &  4  &  136  &  0.3212    &   0.3235 \\ 
\hline
\end{tabular} 
}
\end{table}

Finally, we perform the main task, which is text clustering for the FVnC dataset.  Note that, the ground truth classes of the FVnC dataset are unknown. In order to estimate the number of specific clusters for K-Means clustering method, we take full advantage of DBSCAN algorithm.  In the same way, we get the number of clusters from Algorithm \ref{alg:two}. Specifically, we choose the number of clusters through Silhouette scores with large values corresponding to each pair of parameters $\varepsilon$ and \textit{MinPts}. For FVnC dataset, the optimal values of $\varepsilon$  can be found in Fig.\ref{fig:kdistance}(b). After applying DBSCAN algorithm, the number of clusters actually obtained must be subtracted by 1 for noisy objects (points labeled -1). Before using K-Means algorithm to perform clustering, these noisy objects are removed from FVnC dataset, which leads to the form of the corresponding denoised FVnC datasets. Lastly, we obtain clustering results of FVnC dataset without noise from the K-Means method based on the Silhouette evaluation. Silhouette scores test on original FVnC and denoised FVnC datasets for several different cluster numbers can be examined in Table \ref{table:silhouette_scores}. In case of using $\varepsilon=0.14$ and $\textit{MinPts}=25$, the best obtained Silhouette scores for the original FVnC and denoised FVnC dataset are 0.3359 and 0.3460 respectively. 

\begin{figure}[!htb]
\centering
\includegraphics[width=0.8\textwidth]{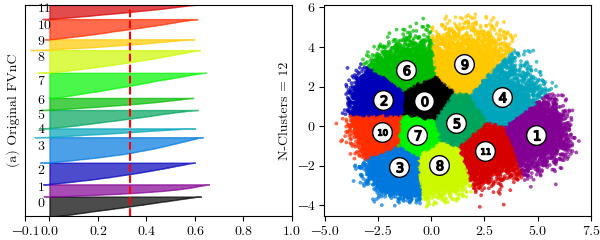}

\includegraphics[width=0.8\textwidth]{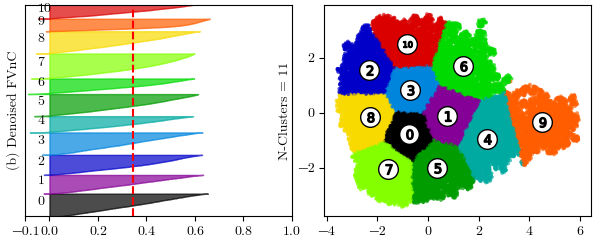}
\caption{Silhouette plots and its clustering visualization} \label{fig:silhouette_FVnC}
\end{figure}

The Silhouette plots and its clustering visualization of the best Silhouette scores can be seen in Figure \ref{fig:silhouette_FVnC}. By comparing the Silhouette plots, we believe that the clustering result on the denoised FVnC dataset (see Fig.\ref{fig:silhouette_FVnC}(b)) is better than the original FVnC dataset (see Fig.\ref{fig:silhouette_FVnC}(a)), because there are some noise points outside the clusters and the size of the silhouette plots of clusters 3, 4 are wide fluctuation in the visualization of Fig.\ref{fig:silhouette_FVnC}(a). With the number of clusters received from the DBSCAN algorithm, the K-Means algorithm gives very good clustering results for denoised FVnC datasets. After all, the clusters obtained from the clustering process are considered as big intents to help us build data for chatbot. For this reason, we can save time and effort and build chatbot faster.

\section{Conclusion}\label{sec:6}
In this work, we research transformer architecture as well as pre-trained language models such as BERT and PhoBERT.  We also cover how to apply PhoBERT to our Facebook Vietnamese conversations dataset (FVnC). Furthermore, we have also built a tool to crawl conversations from a Facebook Messenger Page. After extracting embedding vectors at the final hidden layer, we use the unsupervised learning algorithm K-means and DBSCAN to cluster text data. V-measure score and Silhouette score are used to evaluate the performance of clustering algorithms. A GridSearch algorithm that combines these two clustering evaluations is also proposed to find optimal parameters for the DBSCAN algorithm. The algorithm proposed by us obtained better clustering performance than kneedle algorithm through experimentations based on V-measure scores and Silhouette score on the Sample and FVnC datasets. In addition, we compare the efficiency of the PhoBERT$_{base}$ model in feature extraction for clustering tasks with the other models. PhoBERT$_{base}$ model achieves the best V-measure score on the Sample dataset and wiki dataset. We apply the K-Means clustering method with the number of clusters received from the DBSCAN algorithm to cluster the FVnC dataset. Topics obtained from clustering are similar to intents in building chatbot. From a pre-analysis data screening perspective, clustering results are valuable for building stories in our chatbot. Thanks to the implementation of this clustering, we save a lot of time and effort to build data and storylines for training chatbot. 
\section*{Acknowledgments}
This work was partially supported by Nha Trang University (project TR2020-13-42). The authors thank Hien Thao Le for proofreading our manuscript and fruitful discussions. We also address special thanks to the reviewers for their helpful comments and suggestions.

\bibliographystyle{IEEEtranN}
\bibliography{ref}
\end{document}